\documentclass[letterpaper, 10 pt, conference]{ieeeconf}  

\IEEEoverridecommandlockouts                              

\usepackage{times}
\usepackage{epsfig}
\usepackage{graphicx}
\usepackage{amsmath}
\usepackage{amssymb}
\usepackage{xspace}
\usepackage{booktabs}
\usepackage{mathtools}

\usepackage{amsfonts}

\usepackage{epstopdf}
\usepackage{color}

\usepackage{multirow}
\usepackage{pbox}

\usepackage{amsopn}

\usepackage{subcaption}
\usepackage{multirow}
\usepackage{longtable}
\usepackage{array}
\usepackage{makecell}

\usepackage{tikz}

\usepackage{cellspace}
\usepackage{hyperref}

\makeatletter

\DeclareRobustCommand\onedot{\futurelet\@let@token\@onedot}
\def\@onedot{\ifx\@let@token.\else.\null\fi\xspace}

\def\eg{\emph{e.g}\onedot} 
\def\ie{\emph{i.e}\onedot}

\def\etal{\emph{et al}\onedot}

\newcommand{\PAR}[1]{\vskip4pt \noindent {\bf #1~}}

\definecolor{ubpubColor}{rgb}{0.43, 0.5, 0.5}

\newcommand{\UNPUB}[1]{\textcolor{ubpubColor}{#1}}

\title{\LARGE \bf
EagerMOT: 3D Multi-Object Tracking via Sensor Fusion
}

\author{Aleksandr Kim, Aljo\u{s}a O\u{s}ep and Laura Leal-Taix{\'e}
\thanks{
\scriptsize All authors are with the Technical University of Munich. 
E-mail: {\tt \scriptsize aleksandr.kim@tum.de, aljosa.osep@tum.de, leal.taixe@tum.de}}%
}

\begin{document}

\maketitle
\thispagestyle{empty}
\pagestyle{empty}

\begin{abstract}
Multi-object tracking (MOT) enables mobile robots to perform well-informed motion planning and navigation by localizing surrounding objects in 3D space and time. 
Existing methods rely on depth sensors (\eg, LiDAR) to detect and track targets in 3D space, but only up to a limited sensing range due to the sparsity of the signal. On the other hand, cameras provide a dense and rich visual signal that helps to localize even distant objects, but only in the image domain. 
In this paper, we propose EagerMOT, a simple tracking formulation that eagerly integrates all available object observations from both sensor modalities to obtain a well-informed interpretation of the scene dynamics. 
Using images, we can identify distant incoming objects, while depth estimates allow for precise trajectory localization as soon as objects are within the depth-sensing range. 
With EagerMOT, we achieve state-of-the-art results across several MOT tasks on the KITTI and NuScenes datasets. 
Our code is available at \url{https://github.com/aleksandrkim61/EagerMOT}
\end{abstract}

\section{Introduction}

For safe robot navigation and motion planning, mobile agents need to be aware of surrounding objects and foresee their future states. 
To this end, they need to detect, segment, and -- especially critical in close proximity of the vehicle -- precisely localize objects in 3D space across time. 

As shown by Weng and Kitani~\cite{Weng2020_AB3DMOT}, even a simple method that relies on linear motion models and 3D overlap-driven two-frame data association yields a competitive tracking performance when using a strong LiDAR-based 3D object detector~\cite{Shi19CVPR}.
However, compared to their image-based counterparts, methods that rely on depth sensors are more sensitive to reflective and low-albedo surfaces, and can operate only within a limited sensing range due to the sparsity of the input signal. 
On the other hand, image-based methods leverage a rich visual signal to gain robustness to partial occlusions and localize objects with pixel-precision in the image domain, even when objects are too far away to be localized reliably in 3D space~\cite{Voigtlaender19CVPR, Sharma18ICRA}. However, 3D localization of the surrounding objects is vital in mobile robot scenarios. 

\begin{figure}[ht]
\setlength{\fboxsep}{0.3pt}%
\begin{center}
   \fbox{\includegraphics[width=0.98\linewidth]{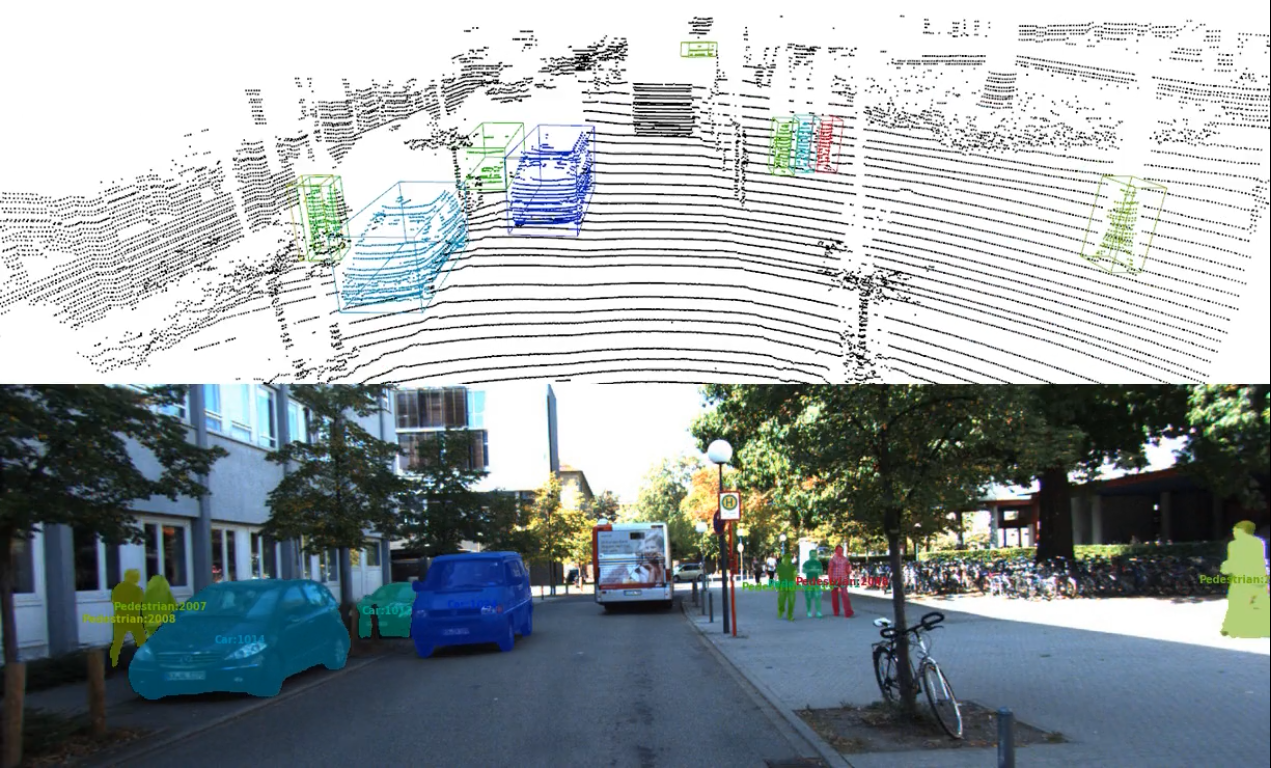}}
\end{center}
\label{fig:qualitative}
\vspace{-7pt}
\caption{Our method eagerly associates different sources of object detection/segmentation information (2D/3D detections, instance segmentation) when available to obtain an, as complete as possible, interpretation of the scene dynamics.}
\end{figure}

In this paper, we present EagerMOT, a simple tracking framework that fuses all available object observations originating from 3D and 2D object detectors, to obtain a well-informed interpretation of the scene dynamics. 
Using cameras, our method identifies and maintains tracks in the image domain, while 3D detections allow for precise 3D trajectory localization as soon as objects enter the LiDAR sensing area. We achieve this via the two-stage association procedure. First, we associate object detections originating from different sensor modalities. Then, we employ a tracking formulation that allows us to update track states even when only partial (either image-based or LiDAR-based) object evidence is available. This way, our EagerMOT is robust to false negatives originating from different sensor modalities and can initialize object tracks before objects enter the depth-sensing range.

Our method is versatile enough to be applied to several different sensory configurations, such as LiDAR combined with a front-facing camera (as used in KITTI~\cite{Geiger12CVPR}), or combined with multiple cameras with non-overlapping view frustums (as employed in NuScenes~\cite{nuscenes2019}). 
With EagerMOT, we establish a new state-of-the-art on the large-scale NuScenes 3D MOT benchmark~\cite{nuscenes2019} and KITTI tracking benchmark~\cite{Geiger12CVPR} for 2D multi-object tracking and segmentation. 

Our method merely assumes a mobile platform with a calibrated sensory setup equipped with a LiDAR and (possibly multiple) camera sensors. Given a pre-trained object detector for both sensor modalities, our method can be easily deployed on any mobile platform without additional training and imposes a minimal additional computational cost at the inference time. 

In summary, \textbf{our contributions} are the following: (i) we propose a simple yet effective multi-stage data association approach that can leverage a variety of different object detectors, originating from potentially different modalities; (ii) we show that our approach can be applied to a variety of MOT tasks (2D/3D MOT and MOTS) and on different sensor configurations; and finally (iii), we perform a thorough analysis of our method, demonstrating through ablation studies the effectiveness of the proposed approach to data association and state-of-the-art results on three different benchmarks.

\section{Related Work}

\PAR{2D MOT.} The majority of the existing vision-based tracking methods rely on recent advances in the field of object detection~\cite{Ren15NIPS, He17ICCV} to detect and track objects in the image domain. 
TrackR-CNN~\cite{Voigtlaender19CVPR} extends Mask R-CNN~\cite{He17ICCV} with 3D convolutional networks {to improve temporal consistency of the detector} and uses object re-identification as a cue for the association. 
Tracktor~\cite{Bergmann19ICCV} re-purposes the regression head of Faster R-CNN~\cite{Ren15NIPS} to follow the targets. Similarly, CenterTrack~\cite{zhou2020tracking} augments the object detector~\cite{zhou2019objects} with an offset-regression head used for cross-frame association.  
Recent trends are going in the direction of end-to-end learning~\cite{xu20cvpr, Frossard18ICRA} and learning to associate using graph neural networks~\cite{braso2020learning, weng20cvpr}.

\PAR{3D MOT.} 
Early methods for LiDAR-based multi-object tracking first perform bottom-up segmentation of LiDAR scans, followed by segment association and track classification~\cite{Teichman11ICRA, Moosmann13ICRA}. 
Due to recent advances in point cloud representation learning~\cite{Qi17CVPR_pointnet, Qi17NIPS} and 3D object detection~\cite{Chen15NIPS, Point-GNN, Shi19CVPR}, LiDAR and stereo-based tracking-by-detection has recently been gaining popularity~\cite{Osep17ICRA, Frossard18ICRA}. 
The recent method by Weng~\etal~\cite{Weng2020_AB3DMOT} proposes a simple yet well-performing 3D MOT method; however, due to its strong reliance on 3D-based detections, it is susceptible to false positives and struggles with bridging longer occlusion gaps. A follow-up method~\cite{chiu2020probabilistic} replaces the intersection-over-union with a Mahalanobis distance-based association measure. 
The recently proposed CenterPoint~\cite{yin2020center} method detects 3D centers of objects and associates them across frames using the predicted velocity vectors. 
In contrast, we propose a method that combines complementary 3D LiDAR object detectors that precisely localize objects in 3D space, and 2D object detectors, that are less susceptible to partial occlusions and remain reliable even when objects are far away from the sensor.

\PAR{Fusion-based methods.} 
Fusing object evidence from 2D and 3D during tracking is an under-explored area. 
O\v{s}ep~\etal~\cite{Osep17ICRA} propose a stereo vision-based approach. At its core, their method uses a tracking state filter that maintains each track's position jointly, in the 3D and the image domain, and can update them using only partial object evidence.
{In contrast, our method treats different sensor modalities independently. We track targets in both domains simultaneously, but we do not explicitly couple their 2D-3D states.} Alternatively, BeyondPixels~\cite{Sharma18ICRA} leverages monocular SLAM to localize tracked objects in 3D space.
MOTSFusion~\cite{luiten19arxiv} fuses optical flow, scene flow, stereo-depth, and 2D object detections to track objects in 3D space. 
Different from that, our method relies only on bounding box object detections obtained from two complementary sensor modalities and scales well across different sensory environments (\eg, single LiDAR and multiple cameras~\cite{nuscenes2019}).
The recently proposed GNN3DMOT~\cite{weng20cvpr} learns to fuse appearance and motion models, independently trained for both images and LiDAR sequences. We compare their method to ours in Sec.~\ref{sec:experimental}.

\section{Method}

\begin{figure*}[ht] 
\vspace{+8pt}
\begin{center}
   \includegraphics[width=1\linewidth]{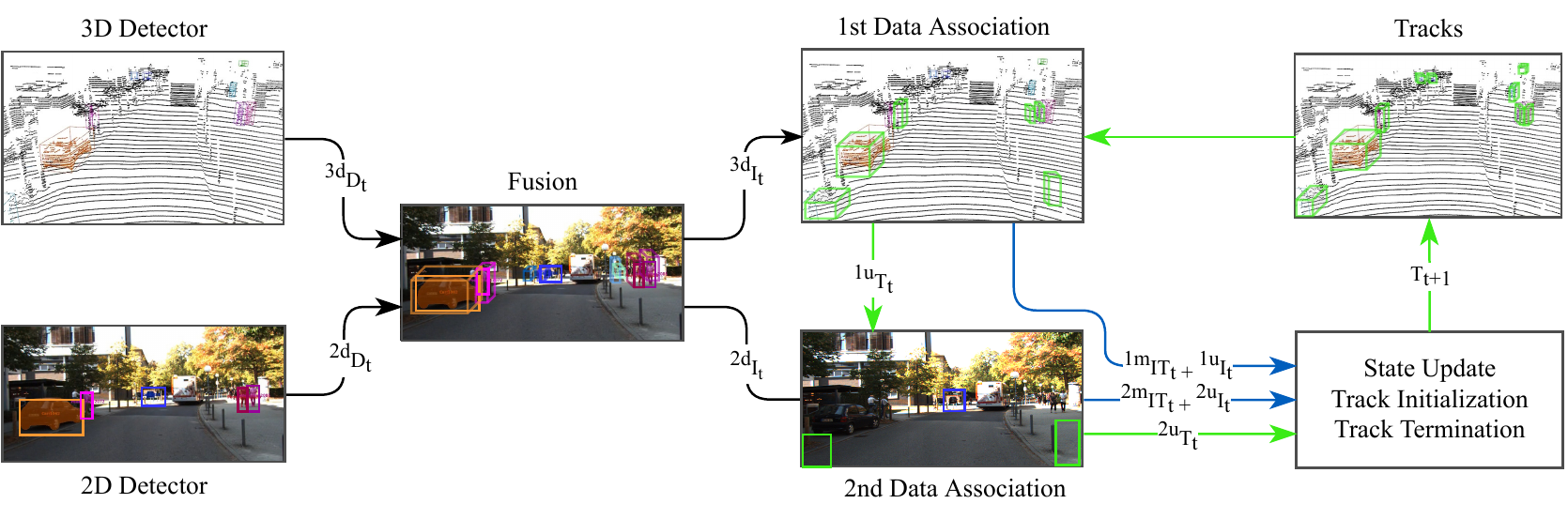}
\end{center}
\vspace{-7pt}
   \caption{A high-level overview of our tracking framework: at the input, we obtain object detections from different sensor modalities, \eg, an image-based detector/segmentation model and a LiDAR/stereo-based 3D object detector. We then fuse these detections into fused object instances, parameterized jointly in 3D and/or 2D space. We then pass them through a two-stage association procedure that allows us to update object tracks, even if detections originating only from one sensor modality are available. 
In the \textit{first stage}, instances with 3D information (with/without 2D information) are matched to existing tracks. In the \textit{second association stage}, unmatched tracks from the previous step ${1u}^T_t$ are matched with instances, localized only in 2D. 
   }
   \label{fig:diagram}
\end{figure*}

Our EagerMOT framework combines complementary 2D, and 3D (\eg, LiDAR) object evidence obtained from pre-trained object detectors. We provide a general overview of our method in Fig.~\ref{fig:diagram}. 
As input at each frame, our method takes a set of 3D bounding box detections $^{3d}D_t$ and a set of 2D detections $^{2d}D_t$. Then, the observation fusion module (i) associates 2D and 3D detections originating from the same objects, (ii) the two-stage data association module associates detections across time, and, based on the available detection information (full 2D+3D, or partial) we update the track states and (iv) we employ a simple track management mechanism to initialize or terminate the tracks. 

This formulation allows all detected objects to be associated to tracks, even if they are not detected either in the image domain or by a 3D sensor. This way, our method can recover from short occlusions and maintain approximate 3D location when one of the detectors fails, and, importantly, we can track far-away objects in the image domain before objects enter the 3D sensing range. Once objects enter the sensing range, we can smoothly initialize a 3D motion model for each track.

\subsection{Fusion} \label{subsection:fusion}

We obtain two sets of object detections at the input, extracted from the input video (2D) and LiDAR (3D) streams. 
LiDAR-based object detections $^{3d}D_t$ are parametrized as 3D object-oriented bounding boxes, while image-based object detections $^{2d}D_t$ are defined by a rectangular 2D bounding box in the image domain. First, we establish a matching between the two sets.

The fusion module performs this task by greedily associating detections in $^{3d}D_t$ to detections in $^{2d}D_t$ based on their 2D overlap in the image domain and produces a set of fused object instances $I_t$=\{$I_t^0$, ..., $I_t^i$\}. 
We define 2D overlap for a pair $^{3d}D_t^i$ and $^{2d}D_t^i$ as the intersection over union (IoU) between the 2D projection of $^{3d}D_t^i$ in the camera image plane and $^{2d}D_t^i$. We note that while different associations criteria could be additionally taken into account (as in, \eg, \cite{Osep17ICRA}), this simple approach was empirically proven to be robust. 

During the greedy association, we sort all possible detection pairings by their overlap in descending order. 
Pairs are considered one-by-one and are combined to form a single fused instance $^{both}I_t^i$ when (i) their overlap is above a threshold $\theta_{fusion}$ and (ii) neither 2D, or 3D detection has been matched yet. 
Fused instances $^{both}I_t \subseteq I_t$ contain information from both modalities: a precise 3D location of the object and its 2D bounding box. Instances may also store additional available information, \eg, a 2D segmentation mask~\cite{Voigtlaender19CVPR}.  
We refer to the remaining (unmatched) detections, which form instances $^{3d}I_t^i \subseteq I_t$ and $^{2d}I_t^i \subseteq I_t$, as \textit{partial observations}, containing information about only one of the two modalities. Note that $^{both}I_t$ $\subseteq$ $^{3d}I_t$ and $^{both}I_t$ $\subseteq$ $^{2d}I_t$.

\PAR{Multi-camera setup.} 
\label{part:multicam}
For scenarios where multiple cameras are available (\eg, in the NuScenes dataset~\cite{nuscenes2019}), we adapt our fusion algorithm as follows. 
In each camera, we perform fusion as explained above; in case a 3D detection is not visible in a particular camera, we consider its overlap with 2D detections in that image plane to be empty. 
After we perform fusion in each 2D plane individually, 3D detections visible through more than one camera might have multiple potential matches. We always associate only one 2D detection with a track and heuristically pick a detection from the view in which the projected 3D bounding box covers the largest area. Other potential pairings from other views are discarded. 

\subsection{Matching}

During each frame $t$, fused instances $I_t$ enter a two-stage matching process to update existing tracks $T_{t}$ with new 3D and/or 2D information.

\PAR{Track parameterization.} As in~\cite{Osep17ICRA}, we maintain 2D and 3D state of tracks $T_t$ in parallel. However, we treat them independently.
We represent the 3D state of a track by a 3D object-oriented bounding box and a positional velocity vector (excluding angular velocity, as in \cite{Weng2020_AB3DMOT}), while a 2D bounding box represents its 2D state. Since we track objects primarily in 3D, a track's confidence score is equal to its 3D state's confidence.
Note that these states do not have to be fully observed for each frame, tracks might be updated using only 3D information $^{3d}T_t \subseteq T_t$, only 2D information $^{2d}T_t \subseteq T_t$, or both $^{both}T_t \subseteq T_t$, $^{both}T_t \subseteq ^{3d}T_t$, $^{both}T_t \subseteq ^{2d}T_t$. 

For tracks $^{3d}T_t$ we additionally maintain a constant-velocity motion model, modeled by a linear Kalman filter. For each new frame $t+1$, existing tracks $^{3d}T_t$ predict their location (an oriented 3D bounding box) in the current frame based on previous observations and velocity estimates.

\PAR{First stage data association.} 
In the first association stage, we match instances detected in 3D with existing tracks using track 3D state information. In particular, we greedily pair detected instances $^{3d}I_t$ with tracks $^{3d}T_{t}$ based on the scaled distance between instances' oriented bounding boxes and tracks' predicted oriented boxes. 
We define the scaled distance for a pair of oriented 3D bounding boxes as the Euclidean distance between them, multiplied by the normalized cosine distance between their orientation vectors: 
\begin{equation}  \label{eq:dist_full}
{
d(B^i, B^j) = \lVert B_{\rho}^i - B_{\rho}^j \rVert * \alpha(B^i, B^j)
},
\end{equation}
\begin{equation}  \label{eq:dist_angle}
{
\alpha(B^i, B^j) = 2 - \cos\langle B_\gamma^i, B_\gamma^j\rangle, \in [1, 2]
},
\end{equation}
where ${B_{\rho}^i} = [x, y, z, h, w, l]$ is a vector containing the 3D location and dimensions of the bounding box and ${B_{\gamma}^i}$ represents the orientation of the box around the vertical axis. 

Compared to planar Euclidean distance, this approach takes into account orientation similarity, which can be informative for non-omnidirectional objects such as vehicles or pedestrians. 
Experimentally, we found this association criterion to be more robust compared to 3D IoU~\cite{Weng2020_AB3DMOT} and Mahalanobis distance that takes the predictive and observational uncertainty into account~\cite{chiu2020probabilistic}), especially in low frame rate scenarios (\eg, in NuScenes dataset~\cite{nuscenes2019}).

Similar to {\subsectionautorefname} {\ref{subsection:fusion}}, best-matching instance-track pairs (below a maximum threshold $\theta_{3d}$) form successful matching tuples $^{1m}IT_t = \{(I_t^i, T_t^j), ...\}$. 
We label the rest as unmatched, \ie, instances $^{1u}I_t$ and tracks $^{1u}T_t$. After this first matching stage, all object instances detected in 3D should be successfully associated with existing tracks or be labeled as unmatched and will not participate in further matching.

\PAR{Second stage data association.} In the second stage, we match detected instances to tracks in the 2D image domain. 
We greedily associate instances {$^{2d}I_t \setminus ^{both}I_t$} to remaining tracks {$^{1u}T_t$ $\cup$ $^{2d}T_t$} based on the 2D IoU criterion. For each instance-track pair, we evaluate the overlap between the instance's 2D bounding box in the current frame and the 2D projection of the track's predicted 3D bounding box or the last observed 2D bounding box in case a 3D prediction is not available (for $^{2d}T_t$). Note that instances that were detected in 3D do not participate in this matching stage even if they were also detected in 2D, \ie, $^{both}I_t$.

This association stage is identical to the first one, except we use here 2D box IoU as the association metric, together with its threshold $\theta_{2d}$. 
Similarly, the output of this stage are a set of matches $^{2m}IT_t = \{(I_t^i, T_t^j), ...\}$, a set of unmatched instances $^{2u}I_t$, and unmatched tracks $^{2u}T_t$. 
In the case of multiple cameras being available, we modify the algorithm as described earlier in {\partautorefname} {\ref{part:multicam}}.

We use a 3D motion model to obtain 2D bounding box predictions in the image domain using a camera projection operation. 
There is not enough 3D evidence to initialize the motion model reliably for certain tracks -- this usually happens for objects observed outside of the LiDAR sensing range. In such scenarios, the apparent bounding box motion is usually negligible, and association can be made purely based on observed 2D boxes. 
Adding a prediction model for the 2D state (as in~\cite{Osep17ICRA, weng20cvpr}) or a (learned) appearance model~\cite{LealTaixe16CVPRW, Voigtlaender19CVPR} could be used to improve the second association stage further and remains our future work.

\PAR{State update.} 
We use matched detected instances to update corresponding tracks with new 3D and/or 2D state information. We simply update the 2D state (top-left and bottom-right bounding box corners) by over-writing the previous state with the newly-detected 2D bounding box. 
We model the 3D state of a track (\ie, object-oriented bounding box parameters) as a multi-variate Gaussian and filter its parameters using a constant-velocity linear Kalman filter (exactly as in~\cite{Weng2020_AB3DMOT}). 
When 3D object detection information is not available (\eg, we have a partial observation providing only a 2D bounding box or a  segmentation mask in the image domain, we only perform the Kalman filter \textit{prediction step} to extrapolate the state.

\subsection{Track lifecycle}

Following AB3DMOT~\cite{Weng2020_AB3DMOT}, we employ a simple set of rules to manage object trajectories and their lifecycle. 
A track is discarded if it has not been updated with any instance (either 3D or 2D) in the last $Age_{max}$ frames. 
As 3D object detectors are usually not as reliable as image-based detectors in terms of precision, a track is considered confirmed if it was associated with an instance in the current frame and has been updated with 2D information in the last $Age_{2d}$ frames. 
Finally, all detected instances $^{2u}I_t$ that were never matched start new tracks.


\section{Experimental Evaluation}  

\begin{table}
\begin{center}
\vspace{+8pt}

\begin{tabular}{l|c|c|c|c}


\toprule
Method & AMOTA & MOTA & Recall & IDs \\
\hline\hline
\textbf{Ours} & \textbf{0.68} & \textbf{0.57} & \textbf{0.73} & 1156 \\
\UNPUB{CenterPoint}~\cite{yin2020center} & 0.65 & 0.54 & 0.68 & \textbf{684} \\
\UNPUB{StanfordIPRL-TRI}~\cite{chiu2020probabilistic} & 0.55 & 0.46 & 0.60 & 950 \\
AB3DMOT~\cite{Weng2020_AB3DMOT} & 0.15 & 0.15 & 0.28 & 9027 \\

\bottomrule
\end{tabular}
\end{center}
\vspace{-7pt}
\caption{Results on the NuScenes 3D MOT benchmark. Methods marked in gray are not yet peer-reviewed.}
\label{tab:nuscenes_test}
\end{table}

\begin{table}
\scriptsize
\begin{center}

\begin{tabular}{c|l|c|c|c|c|c}
\toprule
& Method & Input & sAMOTA & MOTA & MOTP & IDs\\
\hline\hline
\parbox[t]{1mm}{\multirow{6}{*}{\rotatebox[origin=c]{90}{car}}} 
& \textbf{Ours} & 2D+3D & 94.94 & \textbf{96.61} & \textbf{80.00}  & 2 \\
& \textbf{Ours}$^{\dagger}$ & 2D+3D &  \textbf{96.93} & 95.29 & 76.97 & {1} \\
%
%
& GNN3DMOT~\cite{weng20cvpr} & 2D+3D &   93.68 & 84.70 & 79.03 & 10 \\

& mmMOT~\cite{zhang2019robust}  & 2D+3D &  70.61 & 74.07 & 78.16 & 125 \\
& FANTrack~\cite{baser2019fantrack} & 2D+3D &  82.97 & 74.30 & 75.24 &  202 \\

& AB3DMOT$^{\dagger}$~\cite{Weng2020_AB3DMOT} & 3D &  91.78 & 83.35 & 78.43 &  \textbf{0} \\

\hline\hline
\parbox[t]{1mm}{\multirow{3}{*}{\rotatebox[origin=c]{90}{ped.}}} 
%
& \textbf{Ours} & 2D+3D &  \textbf{92.92} & \textbf{93.14} & \textbf{73.22} &  36 \\
& \textbf{Ours}$^{\dagger}$ & 2D+3D &  80.97 & 81.85 &  66.16  & \textbf{0} \\
& AB3DMOT$^{\dagger}$~\cite{Weng2020_AB3DMOT} & 3D &  73.18 & 66.98 & 67.77  & 1 \\
\bottomrule
\end{tabular}
\end{center}
\vspace{-7pt}
\caption{3D MOT evaluation on the KITTI val set (following evaluation protocol by~\cite{Weng2020_AB3DMOT}). Methods marked with $\dagger$ use the Point R-CNN~\cite{Shi19CVPR} 3D object detector. Baseline results taken from~\cite{weng20cvpr}. Note that several methods are only reported results for the \textit{car} class.}
\label{tab:mot3d}
\end{table}
\begin{table*}
\begin{center}
\vspace{+8pt}
\begin{tabular}
{l|c|c|c|c|c|c|c|c|c|c|c|c}

\toprule
 &  & \multicolumn{5}{c|}{car} && \multicolumn{5}{c}{pedestrian}  \\
Method                                 & Inputs             & HOTA  & DetA  & AssA  & MOTA  & IDs          &          & HOTA  & DetA  & AssA  & MOTA  & IDs \\

\hline\hline
\textbf{Ours}                           & 2D+3D (LiDAR)     & \textbf{74.39} & 75.27 & 74.16 & 87.82 & 239          &          & 39.38 & 40.60 & 38.72 & 49.82 & 496 \\
mono3DT~\cite{Hu_2019_ICCV}             & 2D+GPS            & 73.16 & 72.73 & \textbf{74.18} & 84.28 & 379          &          & -- & -- & -- & -- & -- \\
CenterTrack~\cite{zhou2020tracking}     & 2D                & 73.02 & \textbf{75.62} & 71.20 & \textbf{88.83} & 254          &          & 40.35 & \textbf{44.48 }& 36.93 & \textbf{53.84} & 425 \\
SMAT~\cite{gonzalez2020smat}            & 2D                & 71.88 & 72.13 & 72.13 & 83.64 & 198          &          & -- & -- & -- & -- & -- \\
3D-TLSR~\cite{nguyen20203d}             & 2D+3D (stereo)    & --    &  --   & --    & --    & --           &          & \textbf{46.34} & 42.03 & \textbf{51.32} & 53.58 & \textbf{175} \\
Be-Track~\cite{s19020391}               & 2D+3D (LiDAR)     & --    &  --   & --    & --    & --           &          & 43.36 & 39.99 & 47.23 & 50.85 & 199 \\
AB3DMOT~\cite{Weng2020_AB3DMOT}         & 3D (LiDAR)        & 69.81 & 71.06 & 69.06 & 83.49 & \textbf{126}          &          & 35.57 & 32.99 & 38.58 & 38.93 & 259 \\
JRMOT~\cite{shenoi2020jrmot}            & 2D+3D (RGB-D)     & 69.61 & 73.05 & 66.89 & 85.10 & 271          &          & 34.24 & 38.79 & 30.55 & 45.31 & 631 \\
MOTSFusion~\cite{luiten19arxiv}         & 2D+3D (stereo)    & 68.74 & 72.19 & 66.16 & 84.24 & 415          &          & -- & -- & -- & -- & -- \\
MASS~\cite{8782450}                     & 2D                & 68.25 & 72.92 & 64.46 & 84.64 & 353          &          & -- & -- & -- & -- & -- \\
BeyondPixels~\cite{Sharma18ICRA}        & 2D+3D (mono SLAM) & 63.75 & 72.87 & 56.40 & 82.68 & 934          &          & -- & -- & -- & -- & -- \\
mmMOT~\cite{Zhang_2019_ICCV}            & 2D+3D             & 62.05 & 72.29 & 54.02 & 83.23 & 733          &          & -- & -- & -- & -- & -- \\

\bottomrule

\end{tabular}
\end{center}
   \vspace{-7pt}
\caption{Results on the 2D MOT KITTI benchmark. Note: reported methods use different object detectors, \eg our method uses the RRC~\cite{Ren17CVPR} detector for the \textit{car} class, same as MOTSFusion~\cite{luiten19arxiv} and BeyondPixels~\cite{Sharma18ICRA}.}

\label{tab:mot2d}
\end{table*}

\label{sec:experimental}

We evaluate our method using two datasets, KITTI~\cite{Geiger12CVPR} and NuScenes~\cite{nuscenes2019} using four different multi-object tracking benchmarks: (i) NuScenes 3D MOT, (ii) KITTI 3D MOT, (iii) KITTI 2D MOT, and (iv) KITTI MOTS~\cite{Voigtlaender19CVPR}. 
For NuScenes 3D MOT, KITTI 2D MOT, and KITTI MOTS, we use the official benchmarks and compare our method to published and peer-reviewed state-of-the-art methods. 

\PAR{Evaluation measures.} We discuss the results using standard CLEAR-MOT evaluation measures~\cite{Bernardin08JIVP} and focus the discussion on the \textit{multi-object tracking accuracy} (\textit{MOTA}) metric. 
For KITTI 3D MOT, we follow the evaluation setting of \cite{Weng2020_AB3DMOT} and report averaged variants of CLEAR-MOT evaluation measures (\textit{AMOTA} and \textit{AMOTP} stand for averaged \textit{MOTA} and \textit{MOTP}).
For MOTS, we follow the evaluation protocol of~\cite{Voigtlaender19CVPR} and report \textit{multi-object tracking and segmentation accuracy} (\textit{MOTSA}) and \textit{precision} (\textit{MOTSP}). %
On KITTI 2D MOT and MOTS benchmarks we additionally report the recently introduced \textit{higher-order tracking accuracy} (\textit{HOTA}) metric~\cite{luiten20ijcv}\footnote{The official KITTI 2D MOT benchmark switched to HOTA-based evaluation shortly before releasing this paper, therefore we only report benchmark results using this metric.}. HOTA dis-entangles detection and tracking aspects of the task by separately measuring \textit{detection accuracy} (\textit{DetA}) that evaluates detection performance, and \textit{association accuracy} (\textit{AssA}) that evaluates detection association.

\PAR{3D detections.} 
For our final model on NuScenes, we use detections provided by CenterPoint~\cite{yin2020center}. 
On KITTI 3D MOT, we report and compare results obtained using state-of-the-art Point-GNN~\cite{Point-GNN} and Point R-CNN~\cite{Shi19CVPR} (as used by~\cite{Weng2020_AB3DMOT}) 3D object detectors. 
For our model, submitted to the KITTI benchmark, we used Point-GNN~\cite{Point-GNN}.
We do not pre-filter 3D object detections and take all of them as input to our tracking pipeline.

\PAR{2D detections.} 
On NuScenes, we use the Cascade R-CNN~\cite{cai18cascadercnn, mmdetection} object detector, trained on the NuImages~\cite{nuscenes2019} dataset. 
On KITTI, we follow MOTSFusion~\cite{luiten19arxiv} and use 2D detections from RRC~\cite{Ren17CVPR} for \textit{cars} and TrackR-CNN~\cite{Voigtlaender19CVPR} for \textit{pedestrians}. We use thresholds of $0.6$ and $0.9$ for RRC and TrackR-CNN detections, respectively

\subsection{Ablation studies}
\PAR{Data association.} In \autoref{tab:nuscenes_ablation}, we compare different variants of our method, evaluated on the NuScenes validation set. 
The significant difference between \textit{``Full``} ($0.712$ AMOTA) and \textit{``No 2D info``} ($0.651$ AMOTA) highlights the impact of leveraging 2D object detections on the overall performance. 
We note that as we improve the recall ($+0.054$) with our full model, we observe a decrease in AMOTP ($-0.018$), which measures localization precision, averaged over all trajectories. 
This is because our method can leverage 2D object detections and update track states even when 3D detections are not available. In this case, we cannot update the track state using 3D evidence. However, we can still localize objects by performing Kalman filter predictions at the loss of overall 3D localization precision.  

Next, we ablate the impact of our data association function.  
The configuration \textit{``No 2D info; 2D distance``} highlights the performance of a variant that does not use 2D detection information and performs association by simply computing Euclidean distance (on the estimated 2D ground-plane) between the track prediction and detections as an association criterion for the (only) matching stage. %
The variant \textit{``No 2D info; 3D IoU``} is the variant that uses 3D IoU (as in~\cite{Weng2020_AB3DMOT}) as the association metric. As can be seen, our association function is more robust compared to 2D distance ($+0.004$ AMOTA) and 3D IoU ($+0.036$ AMOTA). We conclude that 3D IoU is not suitable for NuScenes due to a significantly lower scan-rate compared to KITTI.

\PAR{Detection sources.} In \autoref{tab:kitti_ablation_source}, we show the impact of detection quality on overall performance. One of the advantages of our method is its flexibility. Unlike other trackers, our framework does not need expensive training and can be easily applied to off-the-shelf detectors. 
As expected, better detectors lead to better tracking performance.

\begin{table}
\begin{center}

\begin{tabular}{l|c|c|c|c}

\toprule
Method & AMOTA & AMOTP & Recall & IDs \\
\hline\hline
\textbf{Full} & \textbf{0.712} & 0.569 & \textbf{0.752} & 899 \\
No 2D info & 0.651 & 0.587 & 0.698 & 864 \\
No 2D; 2D distance & 0.647 & 0.595 & 0.689 & \textbf{783} \\
No 2D; 3D IoU & 0.615 & \textbf{0.658} & 0.692 & 2749 \\

\bottomrule
\end{tabular}
\end{center}

\vspace{-7pt}
\caption{Data association ablation study, performed on the NuScenes 3D MOT val set.}
\label{tab:nuscenes_ablation}
\end{table}

\begin{table}
\scriptsize
\begin{center}
\resizebox{0.48\textwidth}{!}{

\begin{tabular}{l|l|c|c}

\toprule
3D source & 2D source & MOTA car & MOTA ped \\
\hline\hline
Point-GNN~\cite{Shi19CVPR}  & RRC~\cite{Ren17CVPR} + Track-RCNN~\cite{Voigtlaender19CVPR} & 92.5 & \textbf{72.4} \\
Point R-CNN~\cite{Weng2020_AB3DMOT} & RRC~\cite{Ren17CVPR} + Track-RCNN~\cite{Voigtlaender19CVPR} & \textbf{92.7} & 65.6 \\
Point-GNN~\cite{Shi19CVPR} & Cascade R-CNN~\cite{cai18cascadercnn} & 89.0 & 69.5 \\

\bottomrule
\end{tabular}
}
\end{center}
\vspace{-7pt}
\caption{Ablation on the effect of using different object detection sources (KITTI 2D MOT val set).}
\label{tab:kitti_ablation_source}
\end{table}

\subsection{Benchmark results}

\PAR{NuScenes.} 
We report the results obtained using the official NuScenes large-scale tracking benchmark in \autoref{tab:nuscenes_test}. 
In addition to published methods, we include in our analysis the highest-ranking unpublished method~\cite{yin2020center}, marked with \UNPUB{gray}. 
The test set includes $150$ scenes, 80 seconds each. This is a challenging benchmark due to a wide variety of object classes and a low frame rate of 2FPS.

For a fair comparison, we use the same 3D detections as CenterPoint~\cite{yin2020center}. However, we only use 3D bounding box information (and not the predicted velocity vectors). 
The difference in recall supports our assumption that fusing 2D detections helps to bridge occlusions and recover tracks that would otherwise be lost. 

\PAR{KITTI 3D MOT.} \label{part:kitti-results}
In \autoref{tab:mot3d}, we compare our 3D MOT performance to several baselines, as reported in~\cite{weng20cvpr}. Methods marked with $\dagger$ use the Point R-CNN~\cite{Shi19CVPR} 3D object detector. As our method uses the same 3D detections as AB3DMOT, we can conclude that the improvements ($+5.15$ and $+7.79$ sAMOTA for \textit{car} and \textit{pedestrian} classes, respectively) show the merit of our two-stage association procedure that leverages 2D information to improve 3D MOT performance.

\begin{figure}[ht]
\setlength{\fboxsep}{0.3pt}%
\vspace{+8pt}
\begin{center}
    \fbox{\includegraphics[width=0.95\linewidth]{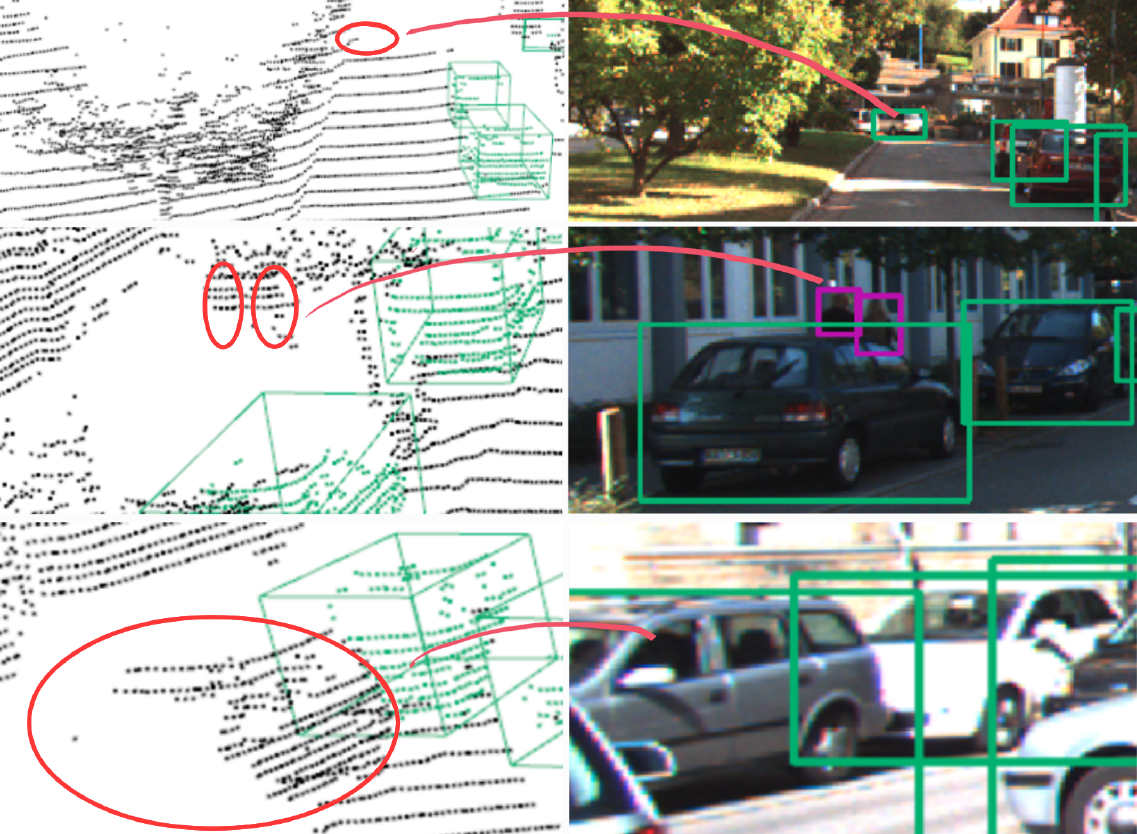}}
\end{center}
\vspace{-7pt}
\caption{Examples of objects overlooked by the 3D detector but recognized by the image-based detector. 
From the top: out of range, partially occluded, detector failure.
}
\label{fig:missed_detections}
\end{figure}

\PAR{KITTI 2D MOT.}
In \autoref{tab:mot2d}, we report 2D MOT results we obtain on the KITTI test set.
Even though we track objects in 3D space, we can report 2D tracking results by projecting 3D bounding boxes to the image plane using camera intrinsics and report minimal axis-aligned 2D bounding boxes that fully enclose those projections as tracks' 2D positions.
Even though we track objects only in 3D, use 2D detections only as a secondary cue and report approximate 2D locations, we achieve state-of-the-art results in terms of the HOTA metric. 
In Fig.~\ref{fig:missed_detections}, we highlight examples where 3D detector fails due to signal sparsity or occlusions; however, we obtain 2D object detections, which we use to update the track states. This example demonstrates that 3D LiDAR and image-based detections are complementary cues for tracking. 
We note that our method performs especially well in terms of the association accuracy ($74.16$ \textit{AssA} on the \textit{car} class), confirming that the secondary association stage does improve not only the detection aspect of the task but also the temporal association.
For the \textit{pedestrian} class, we obtain lower performance compared to current top entries, as we are using weaker object detectors (TrackR-CNN~\cite{Voigtlaender19CVPR}, trained on the non-amodal MOTS dataset) and Point-GNN~\cite{Point-GNN}.

\PAR{KITTI MOTS.}
Multi-object tracking and segmentation (MOTS) extends MOT with pixel-precise localization of object tracks. 
We can easily adapt our MOTS approach by additionally passing segmentation masks from instances to tracks after the data association. 

In \autoref{tab:mots}, we report our MOTS performance on the KITTI test set and compare it to other published methods.
As can be seen, we obtain better results compared to MOTSFusion on both classes ($+1.03$ for \textit{car} and $+3.61$ for \textit{pedestrian} class) despite using the same set of 2D segmentation masks. 
We note, however, that EagerMOT additionally used 3D object detections obtained from the LiDAR stream, while MOTFusion relies on stereo cameras. Our method is applicable to a wide variety of LiDAR-centric sensory configurations, often employed in modern automotive datasets, \eg, NuScenes~\cite{nuscenes2019}, Waymo Open Dataset~\cite{sun2020scalability} and Argoverse~\cite{chang2019argoverse}. 
Moreover, our method runs at $90$ FPS on KITTI (LiDAR + single camera) compared to MOTSFusion at $2$ FPS (stereo cameras).\footnote{Both exclude the time spent on object detection and ego-motion estimation.} 
Finally, our method establishes new state-of-the-art results for both, \textit{car} ($74.66$) and \textit{pedestrian} ($57.65$) classes in terms of HOTA. Furthermore, our method performs especially well in terms of association accuracy (\textit{AssA}), again confirming that using additional sensor observations helps to maintain track consistency.

\begin{table}
\scriptsize
\vspace{+7pt}
\begin{center}
\resizebox{0.48\textwidth}{!}{

\begin{tabular}{c|l|c|c|c|c|c|c}

\toprule
& Method & HOTA  & DetA  & AssA  & sMOTSA & IDs & FPS \\
\hline\hline
\parbox[t]{1mm}{\multirow{4}{*}{\rotatebox[origin=c]{90}{car}}} 
& \textbf{Ours}                                 & \textbf{74.66} & 76.11 & \textbf{73.75} & 74.53 & 458 & \textbf{90}  \\
& MOTSFusion~\cite{luiten19arxiv}      & 73.63 & 75.44 & 72.39 & 74.98 & \textbf{201} & 2  \\
& PointTrack~\cite{xu2020Segment}      & 61.95 & \textbf{79.38} & 48.83 & \textbf{78.50} & 346 & 22  \\
& TrackR-CNN~\cite{Voigtlaender19CVPR} & 56.63 & 69.90 & 46.53 & 66.97 & 692 & 2  \\

\hline\hline
\parbox[t]{1mm}{\multirow{4}{*}{\rotatebox[origin=c]{90}{ped.}}} 
& \textbf{Ours} & \textbf{57.65} & 60.30 & \textbf{56.19} & 58.08 & 270 & \textbf{90}  \\
& MOTSFusion                           & 54.04 & 60.83 & 49.45 & 58.75 & 279 & 2  \\
& PointTrack                           & 54.44 & \textbf{62.29} & 48.08 & \textbf{61.47} & \textbf{176} & 22  \\
& TrackR-CNN                           & 41.93 & 53.75 & 33.84 & 47.31 & 482 & 2  \\

\bottomrule
\end{tabular}
}
\end{center}
\vspace{-7pt}
\caption{Results on the 2D KITTI MOTS benchmark (for \textit{car} and \textit{pedestrian} classes). Note: our method uses the same set of object detections and segmentation masks as MOTSFusion.}
\label{tab:mots}
\end{table}

\subsection{Runtime discussion}

Excluding the time spent on object detection and ego-motion estimation, our Python implementation runs at 4 FPS on NuScenes. 
It is slower (but more accurate) compared to StanfordIPRL-TRI~\cite{chiu2020probabilistic} and AB3DMOT~\cite{Weng2020_AB3DMOT} that only use LiDAR data and run at 10 FPS.

On KITTI, our method runs at 90 FPS because we only have a single camera and do not need to perform multi-camera association. 
We report higher frame rates compared to several 3D MOT methods reported on the KITTI benchmark, including GNN3DMOT (5 FPS), mmMOT (4 FPS), and FANTrack (25 FPS), that also leverage both 2D and 3D input. 

\PAR{Implementation details.}
On KITTI, we use $\theta_{fusion}=0.01$, $\theta_{3d}=0.01$, $\theta_{2d}=0.3$, $Age_{max}=3$, and $Age_{2d}=3$ for both classes.
For 2D MOT evaluation, we report 2D projections of estimated 3D bounding boxes only for confirmed tracks.

On NuScenes, we use $\theta_{fusion}=0.3$, $\theta_{2d}=0.5$, and $Age_{max}=3$ for all seven classes. Other parameters are class-specific:  $\theta_{3d}=(7.5, 1.8, 4.4, 8.15, 7.5, 4.9, 7.5)$, $Age_{2d}=(2, 3, 1, 3, 3, 2, 2)$ for \textit{car, pedestrian, bicycle, bus, motorcycle, trailer, truck}. Additionally, $\theta_{fusion}=0.01$ for \textit{trailer} and $\theta_{fusion}=0.3$ for all other classes.

For 3D MOT evaluation on KITTI and NuScenes, we report estimated 3D boxes for confirmed tracks with their original confidence scores. Estimates for unconfirmed tracks are also reported. However, we halve their scores for each frame for which we do not perform 2D updates.

\section{Conclusion}

We presented a tracking framework that can leverage different sources of object detections originating from varying sensor modalities through a two-stage association procedure.  
Our experimental evaluation reveals that our method performs consistently well across different datasets and tracking tasks and can be used in combination with a variety of different object detectors -- without requiring any additional detector-specific fine-tuning. 
We hope that our framework will serve as a baseline for future research in sensor-fusion-based multi-object tracking.

\footnotesize \PAR{Acknowledgements:} This project was funded by the Humboldt Foundation through the Sofja Kovalevskaja
Award. We thank Paul Voigtlaender for his feedback. 

{\small
\bibliographystyle{ieee}
\bibliography{abbrev_short,refs}
}

\end{document}